\documentclass[10pt,twocolumn,letterpaper]{article}
\usepackage[pagenumbers]{cvpr}
\usepackage[pagebackref=true,breaklinks=true,colorlinks,citecolor=cyan,linkcolor=red,bookmarks=false]{hyperref}
\usepackage{subcaption}
\usepackage{booktabs}

\usepackage{multirow}
\usepackage{animate}
\usepackage{graphicx}
\usepackage{wrapfig}
\usepackage{hyperref}
\usepackage{url}
\newcommand{\myPara}[1]{\noindent\textbf{#1}}
\newcommand{\secref}[1]{Sec.~\ref{#1}}
\newcommand{\tabref}[1]{Tab.~\ref{#1}}
\newcommand{\figref}[1]{Fig.~\ref{#1}}

\usepackage{pifont}%

\newcommand{\cmark}{\ding{51}}%
\newcommand{\xmark}{\ding{55}}%

\title{ROICtrl: Boosting Instance Control for Visual Generation}

\author{Yuchao Gu$^{1,2}$,\; Yipin Zhou$^{2}$,\; Yunfan Ye$^{2}$,\; Yixin Nie$^{2}$,\; Licheng Yu$^{2}$,\; \\ Pingchuan Ma$^{3}$\textbf{,}\; Kevin Qinghong Lin$^{1}$,\; Mike Zheng Shou$^{1}$\thanks{Corresponding Author.}\;\\ \\
$^1$Show Lab, National University of Singapore\quad $^2$GenAI, Meta \quad $^3$MIT\\
\url{https://roictrl.github.io/}
}

\begin{document}
\maketitle

\begin{abstract}
Natural language often struggles to accurately associate positional and attribute information with multiple instances, which limits current text-based visual generation models to simpler compositions featuring only a few dominant instances. To address this limitation, this work enhances diffusion models by introducing regional instance control, where each instance is governed by a bounding box paired with a free-form caption. Previous methods in this area typically rely on implicit position encoding or explicit attention masks to separate regions of interest (ROIs), resulting in either inaccurate coordinate injection or large computational overhead. Inspired by ROI-Align in object detection, we introduce a complementary operation called ROI-Unpool. Together, ROI-Align and ROI-Unpool enable explicit, efficient, and accurate ROI manipulation on high-resolution feature maps for visual generation. Building on ROI-Unpool, we propose ROICtrl, an adapter for pretrained diffusion models that enables precise regional instance control. ROICtrl is compatible with community-finetuned diffusion models, as well as with existing spatial-based add-ons (\eg, ControlNet, T2I-Adapter) and embedding-based add-ons (\eg, IP-Adapter, ED-LoRA), extending their applications to multi-instance generation. Experiments show that ROICtrl achieves superior performance in regional instance control while significantly reducing computational costs.
\end{abstract}

\section{Introduction}

\begin{figure}[!tb]
\centering
\includegraphics[width=\linewidth]{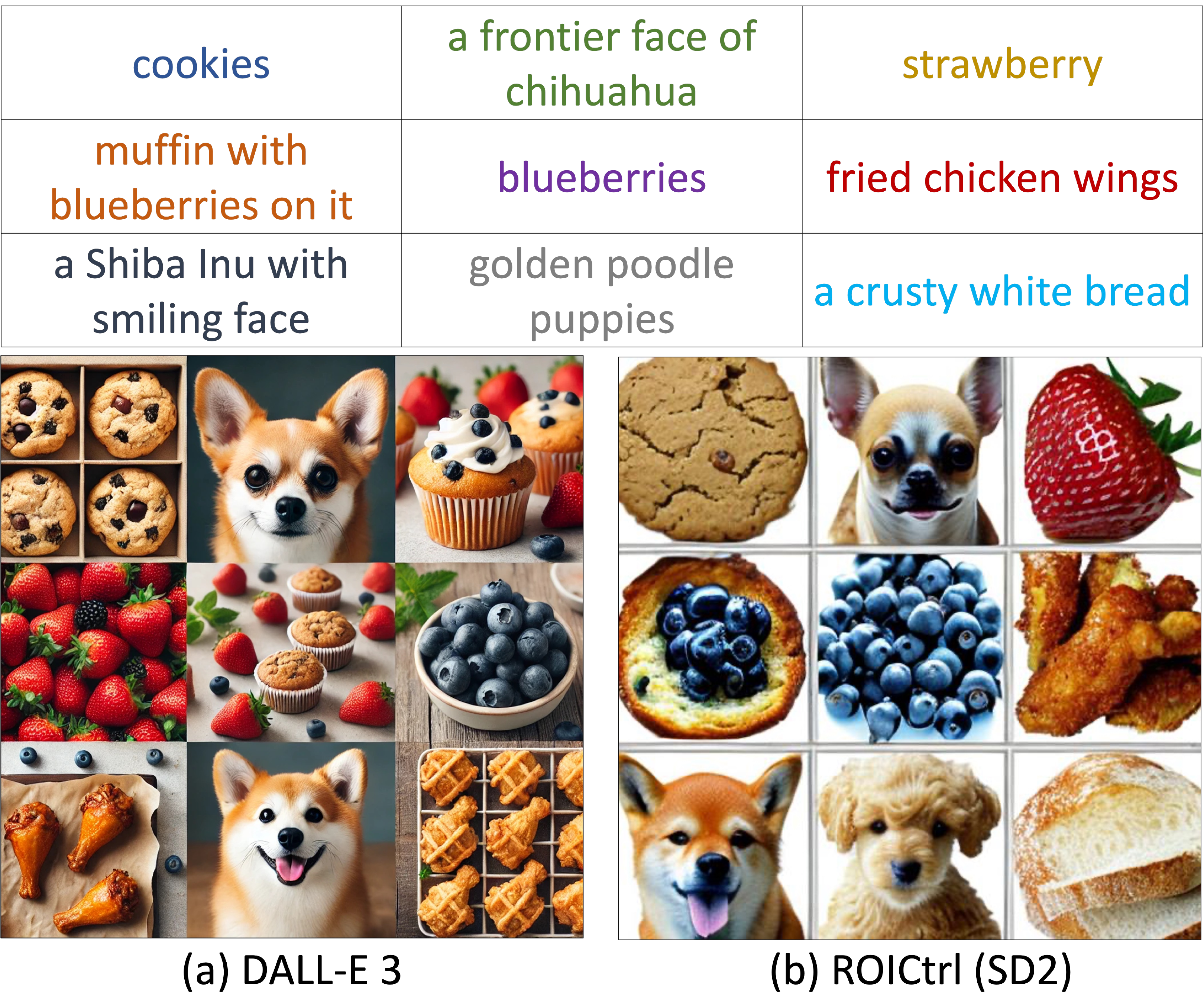}
  \vspace{-.25in}
  \caption{Grid test for instance control. (a) We structure the region positions and instance captions into a single plain caption, then prompt DALL-E 3 to generate a nine-grid image. (b) We apply ROICtrl to generate a nine-grid image based on instance captions.}
  \label{fig:teaser}
  \vspace{-.1in}
\end{figure}

Recent text-based diffusion models have achieved remarkable success in generating images \cite{rombach2022high, dai2023emu, saharia2022photorealistic} and videos \cite{wu2023tune, girdhar2023emu, zhang2023show, ho2022imagen} by scaling up data and computational resources. However, effectively controlling these text-based generative models continues to be a major challenge.
The large information gap between natural language and the visual world complicates the precise description of spatial positions and attributes of multiple instances using language alone, often leading to linguistic ambiguity~\cite{feng2022training}. As a result, current text-based diffusion models are more effective at generating images of simple composition with a limited number of dominant instance. Inspired by the ``\textit{chihuahua or muffin}" grid test \citep{fan2024muffin}, which assesses the fine-grained visual recognition ability of multi-modal large language models, we use instance grids to evaluate the state-of-the-art text-to-image generation system DALL-E 3 \citep{betker2023improving}. As shown in \figref{fig:teaser}, we structure region positions and their corresponding caption into a sentence. However, DALL-E 3 struggle to generate accurate nine-grid results, highlighting the challenge of using natural language alone to solve regional instance control in visual generation.

\begin{figure*}[!tb]
    \centering
    \includegraphics[width=\linewidth]{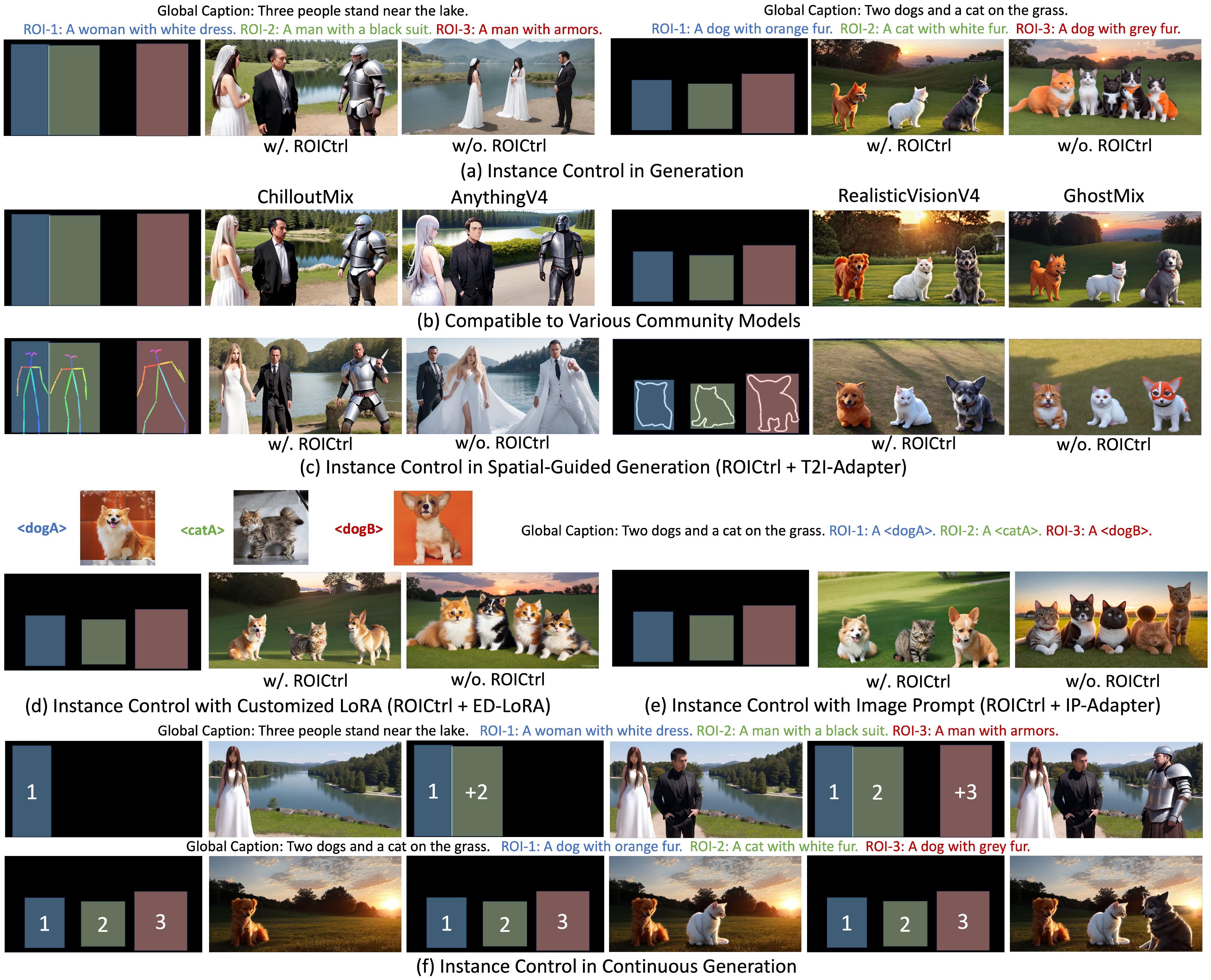}
    \vspace{-.22in}
    \caption{Applications of ROICtrl. A trained ROICtrl adapter can extend existing diffusion models (a) and their community-finetuned versions (b) to multi-instance generation. Additionally, it can collaborate with spatial-based add-ons (c) and embedding-based add-ons (d, e) to offer fine-grained control over spatial or identity information. ROICtrl can also be applied to continuous generation settings (f). \textbf{Due to legal considerations, we do not display customized results involving human identity.}
    }
    \label{fig:application}
    \vspace{-.1in}
\end{figure*}

Much like the evolution in visual recognition, which has transitioned from concentrating on single dominant instances \cite{deng2009imagenet, he2016deep, krizhevsky2012imagenet} (\ie, object classification) to recognizing objects within complex contexts \cite{lin2014microsoft, ren2015faster, he2017mask} (\ie, object detection) through the use of bounding boxes to indicate spatial locations and distinguish instances, visual generation is also shifting towards using bounding boxes to anchor regions of interest (ROIs) for instance control. However, the main difference in ROI processing between visual recognition and visual generation is that \textit{\textbf{visual generation requires handling variable-sized ROIs on high-resolution feature maps}}. For example, in Faster R-CNN~\cite{ren2015faster}, the ROI layer operates on lower-resolution features (\eg, 14$\times$14) with a simple classification head. In contrast, the ROI layer in generative models is applied to higher-resolution features (\eg, 64$\times$64 or 128$\times$128) for finer detail, often using computationally intensive operations like cross- or self-attention.
This has led prior methods to compromise between spatial alignment and computational efficiency in ROI injection.

\begin{figure}[!tb]
\centering
\includegraphics[width=\linewidth]{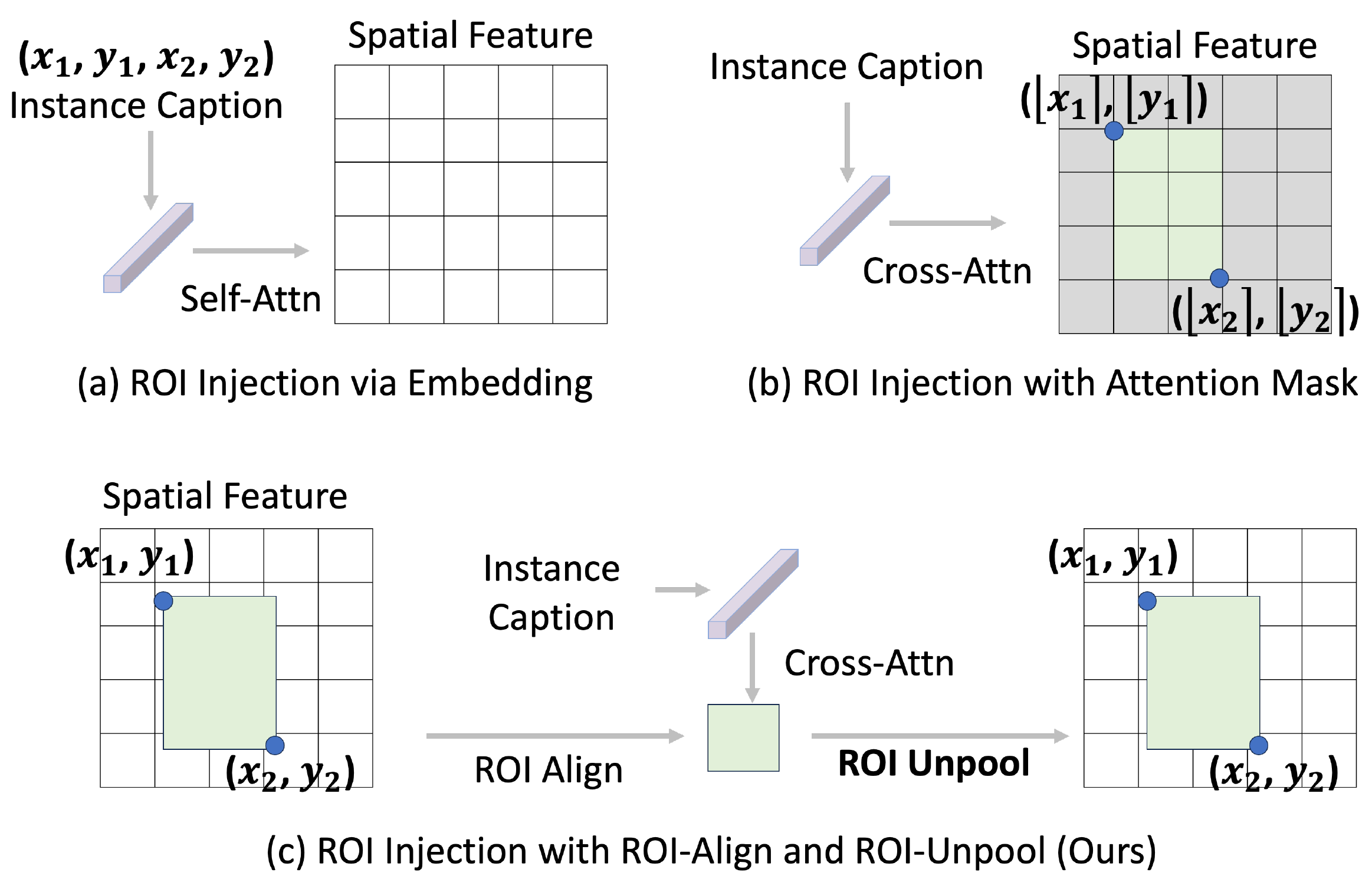}
  \vspace{-.2in}
  \caption{Illustration of different ROI injection designs. $\lfloor\cdot\rceil$ denotes coordinate quantization to the nearest integer.}
  \vspace{-.2in}
  \label{fig:roicomp}
\end{figure}

Prior methods for instance control can be broadly categorized into two approaches:
1) \textbf{Implicit ROI injection via embedding}: As shown in \figref{fig:roicomp}(a), GLIGEN~\cite{li2023gligen} and subsequent works~\cite{feng2024ranni, wang2024boximator} implicitly encode regional information by fusing box coordinate embeddings with instance caption embeddings. Self-attention mechanisms are then used to inject this ROI information into the global feature map. Although implicit ROI injection avoids directly handling variable-sized ROIs, it suffers from severe attribute leakage issues and lower spatial alignment.
2) \textbf{Explicit ROI injection with attention mask}: As shown in \figref{fig:roicomp}(b), MIGC~\cite{zhou2024migc} and Instance Diffusion~\cite{wang2024instancediffusion} use masked cross-attention to isolate each ROI during instance caption injection, achieving better spatial alignment and reducing attribute leakage issues. However, despite the use of masked attention, the computations are still conducted on the full-sized high-resolution feature map, resulting in high computational costs.

In this work, we introduce an effective strategy for instance control in visual generation. Inspired by ROI-Align~\cite{he2017mask} in object detection, we introduce a complementary operation named ROI-Unpool, which restores cropped ROI features to their original position on the high-resolution feature map. As shown in \figref{fig:roicomp}(c), combining ROI-Align and ROI-Unpool allows explicit extraction and processing of ROI features, with computational costs independent of the original feature size.
Building on this operation, we introduce ROICtrl, an adapter that integrates instance control into existing diffusion models. ROICtrl is compatible with existing spatial-based add-ons (\eg, ControlNet~\citep{zhang2023adding} and T2I-Adapter~\cite{mou2024t2i}) and embedding-based add-ons (\eg, IP-Adapter~\cite{ye2023ip} and ED-LoRA~\cite{gu2024mix}), expanding their application for multi-instance generation (as shown in \figref{fig:application}).

In evaluating instance control, we find that previous benchmarks are limited to template-based captions, focusing on specific attributes like color, as summarized in \tabref{tab:bench_comp}. However, users often prefer free-form descriptions to capture broader attributes.
To address this gap, we introduce ROICtrl-Bench, a benchmark specifically designed to evaluate both template-based and free-form instance captions. By leveraging the strong open-domain recognition abilities of multi-modal large language models, ROICtrl-Bench provides a more comprehensive assessment of instance control.

Our contributions are summarized as follows:
\begin{itemize}
\item We introduce ROI-Unpool, an operation that facilitates efficient and accurate ROI injection for visual generation.
\item We propose ROICtrl, an adapter that is compatible with existing diffusion models and their add-ons, expanding their applications in multi-instance generation.
\item We introduce ROICtrl-Bench, a comprehensive benchmark for evaluating instance control capabilities. ROICtrl achieves state-of-the-art performance and improved efficiency on ROICtrl-Bench, as well as on two existing benchmarks (InstDiff-Bench~\cite{wang2024instancediffusion} and MIG-Bench~\cite{zhou2024migc}).
\end{itemize}
\section{Related Work}

\subsection{Controllable Visual Generation}
While text-to-image and text-to-video diffusion models achieve high generation quality, they are limited by language alone in capturing fine-grained spatial(-temporal) and identity details. To address this, researchers have introduced visual conditions to enhance controllability: spatial control for precise layouts (\eg, ControlNet~\cite{zhang2023adding}, T2I-Adapter~\cite{mou2024t2i}), embedding control for detailed identity (\eg, ED-LoRA~\cite{gu2024mix}, IP-Adapter~\cite{ye2023ip}), and trajectory control for fine-grained motion (\eg, VideoSwap~\cite{gu2024videoswap}, MotionCtrl~\cite{wang2024motionctrl}). However, these controls lack explicit instance separation, leading to severe attribute leakage issues in multi-instance generation, as shown in \figref{fig:application}.

\begin{table}[!tb]
\caption{Comparison of existing instance control benchmarks. Previous benchmarks mainly focus on template-based instance captions, while ROICtrl-Bench covers both template-based and free-form instance captions for comprehensive evaluation.}
\vspace{-.12in}
\label{tab:bench_comp}
\resizebox{\linewidth}{!}{\begin{tabular}{lcccc}\toprule
\multirow{2}{*}{\textbf{Benchmarks}} & \multicolumn{2}{c}{\textbf{In-Distribution}} & \multicolumn{2}{c}{\textbf{Out-of-Distribution}} \\
& Template Cap.  & Free-Form Cap. & Template Cap.      & Free-Form Cap.      \\\midrule
GLIGEN-Bench~\cite{li2023gligen} & \cmark &  &  &  \\
MIG-Bench~\cite{zhou2024migc} &  &  & \cmark &  \\
InstDiff-Bench~\cite{wang2024instancediffusion} & \cmark &  & \cmark &  \\\midrule
ROICtrl-Bench  & \cmark & \cmark & \cmark & \cmark \\\bottomrule
\end{tabular}}

\vspace{-.1in}
\end{table}

\subsection{Instance Control in Visual Generation}
Unlike the above controls that enable fine-grained visual alignment, instance control is designed to separate different instances, allowing for independent control of each instance while preventing attribute leakage between them. This approach is often associated with bounding-box, layout, or region control. We group all these types under the term ``\textit{Instance Control}" and outline the main methods below.

\myPara{Training-Free Instance Control.}
Training-free instance control~\cite{xie2023boxdiff, phung2024grounded, he2023localized, chen2024training, kim2023dense} primarily manipulates the attention map in diffusion models during inference, inspired by the finding that cross-attention conveys layout information~\cite{hertz2022prompt}. The core idea is to enhance the influence of nouns on their corresponding regions using techniques such as attention modulation~\cite{kim2023dense} or latent optimization~\cite{xie2023boxdiff, chefer2023attend}. While this approach allows for some degree of instance control, it often involves a trade-off between image quality and spatial alignment, as well as increased computational costs and reduced flexibility during inference.

\myPara{Training-Based Instance Adapter.}
Training-based instance adapters aim to learn instance control from data and can be categorized into implicit and explicit injection methods based on how they incorporate instance information.
Implicit injection~\cite{li2023gligen,wang2024boximator,feng2024ranni} encodes bounding box coordinates as positional embeddings, which are then fused with instance caption embeddings. A gated self-attention mechanism injects the instance embedding into spatial features. While this approach avoids directly handling variable-sized ROIs, it suffers from lower spatial alignment and significant attribute leakage issues.
In contrast, explicit injection isolates the target region during instance caption injection, preventing attribute leakage between instances. Previous works~\citep{zhou2024migc,nie2024compositional,wang2024instancediffusion} adopt attention masks to zero out unrelated regions; however, this approach results in substantial redundant computation and coordinate quantization errors.

To address these limitations, we introduce ROI-Unpool for explicit, efficient, and accurate ROI injection.

\subsection{Instance Recognition in Visual Understanding}

Early research in visual recognition focus on object classification tasks~\cite{deng2009imagenet, he2016deep, krizhevsky2012imagenet}, where each image typically contains a single prominent object (\eg, ImageNet~\cite{deng2009imagenet}). As the field progressed, attention shift to detecting multiple objects in context (\eg, MS-COCO~\cite{lin2014microsoft}), with bounding boxes used to locate individual objects.
The representative approach in object detection is the two-stage detector~\cite{ren2015faster, he2017mask, lin2017feature}, which employs ROI-Pool/ROI-Align to parallelize feature extraction from varied-sized ROIs, enabling efficient ROI processing.
Motivated by this line of research, we explore effective ROI operations for visual generation in this work.
\begin{figure}[!tb]
\centering
\includegraphics[width=0.95\linewidth]{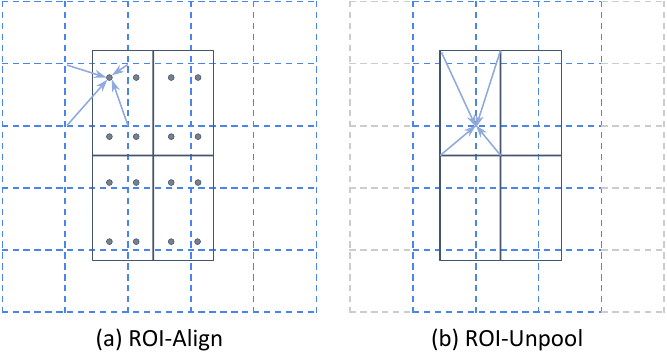}
  \vspace{-.1in}
  \caption{Illustration of ROI-Unpool. The dashed grid represents the spatial features, while the solid grid represents the ROI features. Similar to ROI-Align~\cite{he2017mask}, ROI-Unpool avoids coordinate quantization during computation.}
  \label{fig:roiunpool}
  \vspace{-.22in}
\end{figure}

\section{Methodology}

In this section, we first present the problem formulation of multi-instance generation in \secref{sec:formulation}. Next, we introduce the basic operation, ROI-Unpool, in \secref{sec:roiunpool}. Building on ROI-Unpool, we then describe the design of the ROICtrl adapter in \secref{sec:roictrl}. Finally, we discuss the applications of ROICtrl in \secref{sec:application}.

\subsection{Problem Formulation} 
\label{sec:formulation}

Multi-instance generation is defined as using a global caption
$p_g$ to describe the whole image, along with $n$ instance caption $p_r=\{p_{r_i}\}_{i=1}^n$ and their corresponding bounding box coordinates $c_r=\{c_{r_i}\}_{i=1}^n$ to describe each instance. 
Since the original diffusion model relies solely on the global caption $p_g$ for control, our goal is to develop an adapter that can incorporate region-specific information ($p_r$, $c_r$) into a pretrained diffusion model.
An effective instance adapter should achieve strong spatial alignment and regional text alignment. Beyond these core requirements, the following additional criteria are crucial for real-world applications:

\begin{figure*}[!tb]
    \centering
\includegraphics[width=\linewidth]{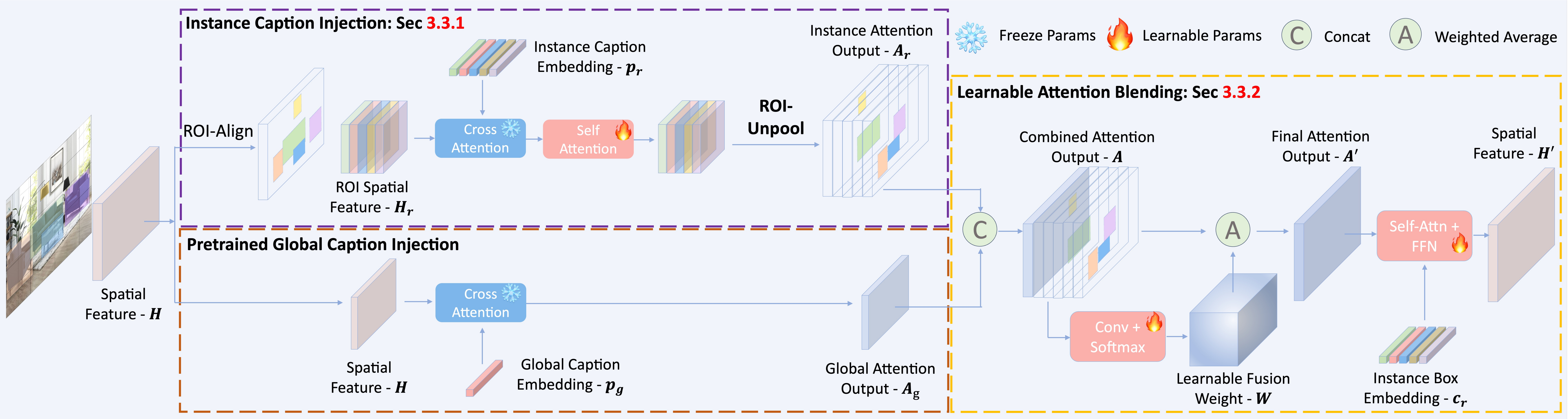}
    \vspace{-.3in}
    \caption{Detailed structure of ROICtrl. In parallel with the pretrained global caption injection, we introduce an additional instance caption injection. The global attention output and instance attention output are then fused using learnable blending.}
\label{fig:pipeline}
\vspace{-.1in}
\end{figure*}

\textbf{1) Free-Form Instance Captions.} Instance captions should be in free-form text, providing flexibility similar to the global caption rather than relying solely on template-based captions~\cite{zhou2024migc, li2023gligen}.

\textbf{2) Compatibility with Fine-Grained Controls.} Since bounding boxes provide only coarse spatial cues for distinguishing instances, the instance adapter should be compatible with existing add-ons that offer fine-grained control, such as spatial-based add-ons (\eg, ControlNet~\cite{zhang2023adding}, T2I-Adapter~\cite{mou2024t2i}) or embedding-based add-ons (\eg, ED-LoRA~\cite{gu2024mix}, IP-Adapter~\cite{ye2023ip}).

\subsection{ROI-Unpool}
\label{sec:roiunpool}

In contrast to object detection, which feeds extracted ROI features into a simple classification head to category prediction, visual generation requires the processed ROI features to be ``\textit{pasted back}" at their original coordinates on the spatial feature map to allow further decoding of fine-grained details.
To achieve this, existing methods~\cite{nie2024compositional, zhou2024migc, wang2024instancediffusion} for instance control primarily use a masked attention mechanism that zeros out unrelated regions during instance caption injection. This approach keeps each ROI at its original spatial coordinates, bypassing the difficulties of “\textit{pasting back}” varied-sized ROIs.
However, the masked attention mechanism introduces significant redundant computation outside the ROI, which is costly on high-resolution feature maps in visual generation. Additionally, coordinate quantization errors during mask creation reduce spatial alignment.

In this work, we address the challenges of ``\textit{pasting back}" varied-sized ROIs onto their original coordinates in the spatial feature map by introducing ROI-Unpool. ROI-Unpool complements ROI-Align~\cite{he2017mask}, enabling explicit ROI operations for visual generation. Specifically, as shown in \figref{fig:roiunpool}, ROI-Align computes ROI features by bilinearly resampling from the four nearest grid points in the original spatial feature map, whereas ROI-Unpool reconstructs the spatial features using the four nearest grid points from the ROI feature. For border regions without all four sample points, we compute partial values from available points. Positions that do not correspond to the ROI region are left empty.

\subsection{ROICtrl}
\label{sec:roictrl}

Building on ROI-Unpool, we introduce ROICtrl, an adapter designed to integrate instance captions into diffusion models. As illustrated in \figref{fig:pipeline}, ROICtrl adds an instance caption injection parallel to the pretrained global caption injection. The global attention output and instance attention output are then fused through a learnable blending mechanism.
Formally, when provided with the global caption $p_g$, $n$ region captions $p_r$ and their coordinates $c_r$, the goal of ROICtrl is to inject the given conditions into the spatial feature $\mathbf{H}$, resulting in
$\mathbf{H}' = \text{ROICtrl}\left( \mathbf{H} \,\big|\, p_g, \{ p_{r_i}, c_{r_i} \}_{i=1}^n \right)
$.

\begin{table*}[!bt]
    \caption{Quantitative comparison with prior works on MIG-Bench~\cite{zhou2024migc}, InstDiff-Bench~\cite{wang2024instancediffusion}, and the proposed ROICtrl-Bench.}
    \label{tab:quantitative}
    \vspace{-.15in}
    \centering
    \begin{subtable}[t]{0.50\linewidth}
        \centering
        \resizebox{\linewidth}{!}{\begin{tabular}{l|cccccc|cccccc}
        \toprule
        \textbf{Method} & \multicolumn{6}{c|}{\textbf{mIoU}} & \multicolumn{6}{c}{\textbf{Instance Success Rate (\%)}} \\\midrule
        Level (\# Instances) & L2 & L3 & L4 & L5 & L6 & AVG & L2 & L3 & L4 & L5 & L6 & AVG \\\midrule
        GLIGEN~\cite{li2023gligen} & 0.37 & 0.29 & 0.253 & 0.26 & 0.26 & 0.27 & 0.42 & 0.32 & 0.27 & 0.27 & 0.28 & 0.30 \\
        MIGC~\cite{zhou2024migc} & 0.64 & 0.58 & 0.57 & 0.54 & 0.57 & 0.56 & 0.74 & 0.67 & 0.67 & 0.63 & 0.66 & 0.66 \\
        Instance Diffusion~\cite{wang2024instancediffusion} & 0.52 & 0.48 & 0.50 & 0.42 & 0.42 & 0.46 & 0.58 & 0.52 & 0.55 & 0.47 & 0.47 & 0.51 \\\midrule
        ROICtrl (Ours) & \textbf{0.78} & \textbf{0.72} & \textbf{0.67} & \textbf{0.61} & \textbf{0.64} & \textbf{0.66} & \textbf{0.85} & \textbf{0.79} & \textbf{0.74} & \textbf{0.67} & \textbf{0.70} & \textbf{0.73} \\ \bottomrule                 
        \end{tabular}}
        \caption{Quantitative evaluation on MIG-Bench~\cite{zhou2024migc}.  MIG-Bench uses Grounding-DINO~\cite{liu2023grounding} mIoU to measure spatial alignment and assesses regional text alignment within the color space.}
        \label{tab:migbench}
    \end{subtable}
    \hfill
    \begin{subtable}[t]{0.49\linewidth}
        \centering
        \resizebox{\linewidth}{!}{\begin{tabular}{l|cccccc|cc|cc}
        \toprule
        \multicolumn{1}{c|}{} & \multicolumn{6}{c|}{\textbf{Location}} & \multicolumn{4}{c}{\textbf{Attribute}} \\\cmidrule{2-11}
        \multicolumn{1}{l|}{\multirow{-2}{*}{\textbf{Method}}} & AP        & $\text{AP}_{50}$      & $\text{AP}_{s}$       & $\text{AP}_{m}$       & $\text{AP}_{l}$       & AR &$\text{Acc}_{\text{color}}$ & $\text{CLIP}_{\text{color}}$ & $\text{Acc}_{\text{texture}}$  & $\text{CLIP}_{\text{texture}}$ \\ \midrule
        \textcolor{gray}{Upper bound (real images)} & \textcolor{gray}{48.4} & \textcolor{gray}{65.2} & \textcolor{gray}{30.9} & \textcolor{gray}{53.3} & \textcolor{gray}{64.8} & \textcolor{gray}{67.8} &
        - & - & - & - \\ \midrule
        GLIGEN~\cite{li2023gligen} & 24.1 & 42.6 & 3.1 & 22.2 & 49.0 & 35.9 & 26.3 & 0.212 & 17.7 & 0.208 \\
        MIGC~\cite{zhou2024migc} & 22.4 & 41.5 & 2.1 & 20.1 & 46.8 & 32.8 & 53.8 & 0.243 & 24.3 & 0.215 \\
        Instance Diffusion~\cite{wang2024instancediffusion} & 40.1 & 57.2 & 10.4 & \textbf{49.4} & \textbf{67.1} & 53.2 & 55.2 & 0.243 & 26.1 & 0.222 \\\midrule
        ROICtrl (Ours) & \textbf{41.0} & \textbf{63.5} & \textbf{16.3} & 46.5 & 65.7 & \textbf{54.1} & \textbf{62.3} & \textbf{0.256} & \textbf{29.3} & \textbf{0.227}\\\bottomrule
        \end{tabular}}
        \caption{Quantitative evaluation on InstDiff-Bench~\cite{wang2024instancediffusion}. InstDiff-Bench evaluates spatial alignment using YOLO-Det~\cite{Jocher_YOLO_by_Ultralytics_2023} Average Precision (AP) and assesses regional text alignment based on color and texture using CLIP score~\cite{radford2021learning}.}
        \label{tab:instdiff_bench}
    \end{subtable}
    \hfill
    \begin{subtable}[t]{\linewidth}
        \centering
        \resizebox{0.78\linewidth}{!}{\begin{tabular}{l|cc|cc|cc|cc|cc}
        \toprule
        & \multicolumn{2}{c|}{\textbf{T1: Subject}} &
          \multicolumn{2}{c|}{\textbf{T2: Subject*}} &
          \multicolumn{2}{c|}{\textbf{T3: Subject + Attribute}} & \multicolumn{2}{c|}{\textbf{T4: Subject + Attribute*}} & \multicolumn{2}{c}{\textbf{AVG}} \\\cmidrule{2-11}
         \multirow{-2}{*}{\textbf{Method}} & mIoU & Acc (\%) &
          mIoU & Acc (\%) &
          mIoU & Acc (\%) &
          mIoU & Acc (\%) &
          mIoU & Acc (\%) \\\midrule
        Upper Bound (real images) & 0.797 & 72.5 & - & - & 0.797 & 66.4 & - & - & - & - \\\midrule
        GLIGEN~\cite{li2023gligen} & 0.579 & 59.1 & 0.474 & 43.3 & 0.546 & 16.3 & 0.548 & 1.90 & 0.537 & 30.2 \\
        MIGC~\cite{zhou2024migc} & 0.521 & 61.9 & 0.442 & 47.6 & 0.498 & 33.7 & 0.498 & 12.3 & 0.490 & 38.9 \\
        Instance Diffusion~\cite{wang2024instancediffusion} & 0.673 & 66.5 & \textbf{0.562} & \textbf{53.5} & 0.634 & 39.4 & 0.559 & 23.0 & 0.607 & 45.6 \\\midrule
        ROICtrl (Ours) & \textbf{0.692} & \textbf{68.9} & 0.557 & 50.9 & \textbf{0.688} & \textbf{47.3} & \textbf{0.669} & \textbf{27.8} & \textbf{0.652} & \textbf{48.7}\\\bottomrule
        \end{tabular}}
        \caption{Quantitative evaluation on the proposed ROICtrl-Bench. We assess spatial alignment using YOLO-World~\cite{cheng2024yolo} mIoU and evaluate regional text alignment with MiniCPM-V 2.6~\cite{yao2024minicpm}. Tracks 1 and 2 examine template-based instance caption, while tracks 3 and 4 evaluate free-form instance caption. * denote out-of-distribution caption rewritten by GPT-4~\cite{achiam2023gpt}.}
    \label{tab:coco_region50K}
    \end{subtable}
    \vspace{-.2in}
\end{table*}

\begin{figure*}[!tb]
    \centering
\includegraphics[width=0.95\linewidth]{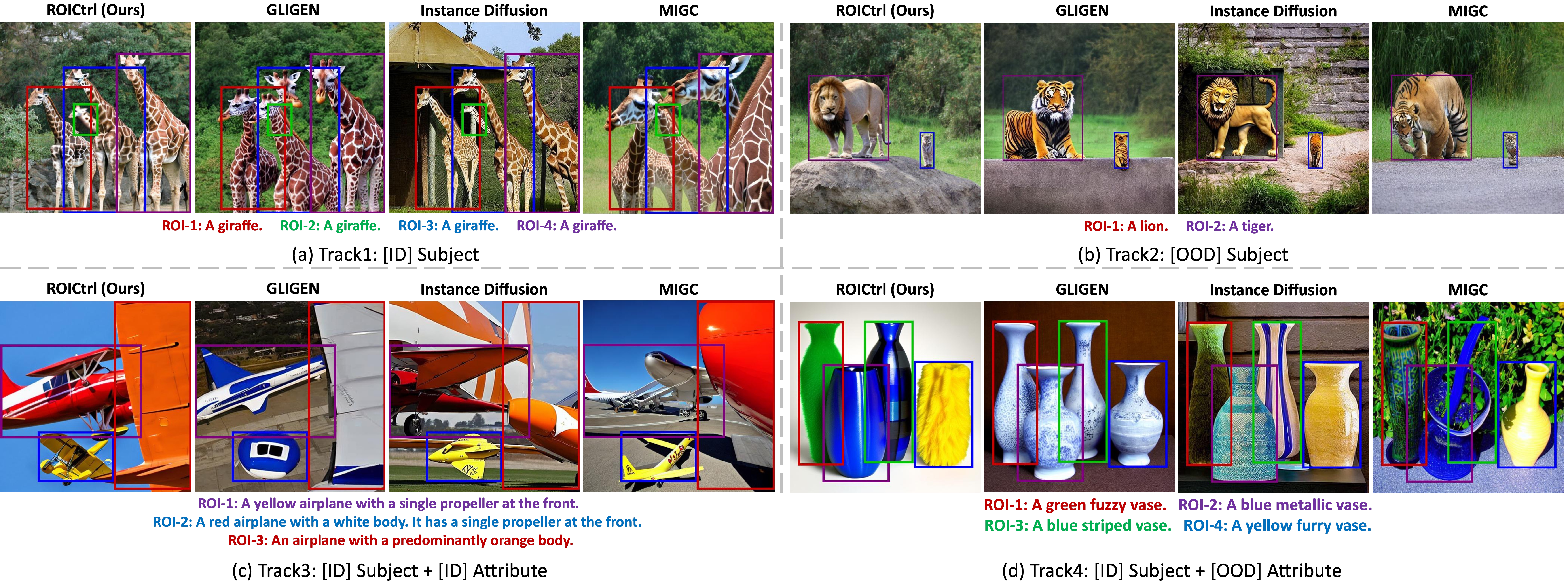}
    \vspace{-.15in}
    \caption{Qualitative comparison on ROICtrl-Bench. Track 1 and 2 examine template-based instance caption, while track 3 and 4 evaluate free-form instance caption. [ID] denotes in-distribution caption derived from real dataset, and [OOD] denotes out-of-distribution caption generated by GPT-4~\cite{achiam2023gpt}.}
    \vspace{-.2in}
\label{fig:qualitative_comp}
\end{figure*}

\subsubsection{Instance Caption Injection}

In ROICtrl, the global caption describes the overall composition and background of the image, while the instance caption provides specific details of each instance.
As shown in \figref{fig:pipeline}, we use the pretrained cross-attention from the diffusion model to generate the global attention output $\mathbf{A}_g \in \mathbb{R}^{b \times c \times h \times w}$. For instance caption injection, we first extract the ROI feature $\mathbf{H}_r \in \mathbb{R}^{b \times n \times c \times r \times r}$ from the spatial feature $\mathbf{H} \in \mathbb{R}^{b \times c \times h \times w}$ using ROI-Align~\cite{he2017mask}, where $r \times r$ represents the ROI feature size, $n$ is the number of ROIs, and $b$, $c$, $h$, and $w$ represent the batch size, channels, height, and width of the spatial feature, respectively.

We then reuse the pretrained cross-attention to inject the instance captions into each ROI feature. Unlike previous methods in \cite{li2023gligen, zhou2024migc, wang2024instancediffusion}, we do not use any additional learnable modules for instance caption injection. This strategy preserves the pretrained model’s knowledge and ensures compatibility with embedding-based add-ons, such as ED-LoRA~\cite{gu2024mix} and IP-Adapter~\cite{ye2023ip}.
Since the pretrained cross-attention is optimized for the original spatial resolution, directly applying it to ROI features may introduce artifacts and misalignment. To address this, we introduce ROI self-attention to refine ROI feature. Finally, ROI-Unpool places the refined ROI features back at their original positions in the feature map, producing the instance attention output $\mathbf{A}_r \in \mathbb{R}^{b \times n \times c \times h \times w}$.

\subsubsection{Learnable Attention Blending}
\label{sec:learnblend}
Given the global attention output $\mathbf{A}_g \in \mathbb{R}^{b \times 1 \times c \times h \times w}$ and the instance attention output $\mathbf{A}_r \in \mathbb{R}^{b \times n \times c \times h \times w}$, we first concatenate them along the ROI axis to form a combined attention output $\mathbf{A} \in \mathbb{R}^{b \times (n+1) \times c \times h \times w}$.

The goal of learnable attention blending is to dynamically reweight the $(n+1)$ attention outputs at each spatial location ($h$, $w$).
To achieve this, we compute the learnable fusion weight $\mathbf{W} \in \mathbb{R}^{b \times n \times 1 \times h \times w}$ by applying a $1\times 1$ convolution to reduce the channel dimension, followed by a softmax function applied across the ROI axis. We then use these weights to perform a weighted fusion of $\mathbf{A}$, which produces the final attention output $\mathbf{A'} \in \mathbb{R}^{b \times c \times h \times w}$.

After obtaining the final attention output  $\mathbf{A'}$, we incorporate instance box embeddings as guidance for occlusion scenario, inspired by GLIGEN~\cite{li2023gligen}. Unlike GLIGEN, we use only box embeddings without instance caption embeddings to prevent attribute leakage. The explicit use of box coordinate conditioning enhances the objectness in the corresponding regions, leading to improved spatial alignment.

\subsubsection{Training Objective}
\label{sec:objective}
ROICtrl is optimized using the standard diffusion loss:
{\setlength{\abovedisplayskip}{1pt}
\setlength{\belowdisplayskip}{1pt}
\footnotesize
\begin{equation}
\mathcal{L}_{\text{LDM}} = \mathbb{E}_{z, \epsilon \sim \mathcal{N}(0, I), t}\left[ \left| \epsilon - \epsilon_\theta(z_t, t, \phi(p_g, p_r, c_r)) \right|_2^2 \right], \label{eq
} \end{equation}}where $\phi$ denotes the learnable parameters of ROICtrl. An additional regularization is applied to the learnable fusion weight $\mathbf{W}$ to reduce the influence of the global attention output within the ROI and facilitate alignment with the instance caption. This regularization term is defined as:
$\mathcal{L}_{\text{reg}} = \frac{\left| \mathbf{M} \odot \mathbf{W}_{:,1,:,:} \right|_1}{\left| \mathbf{M} \right|_1}, $
where $\mathbf{M} \in \{0,1\}^{b \times 1 \times h \times w}$ is a mask identifying the foreground area containing the ROI, and $\mathbf{W}_{:,1,:,:}$ denotes the fusion weight of the global attention output.
The final objective function combines the standard diffusion loss with the regularization term:
 $\mathcal{L} = \mathcal{L}_{\text{LDM}} + \alpha \mathcal{L}_{\text{reg}},$
where $\alpha = 0.01$ throughout our experiments.

\subsection{Application}
\label{sec:application}

In this section, we discuss the applications of ROICtrl.

\myPara{Instance Control.} Without any additional add-ons, ROICtrl alone can be used to control the instance in complex compositions, with each instance can be described with free-form text, as demonstrated in \figref{fig:application}(a).

\myPara{Compatible with Various Community Models.} Once ROICtrl is trained on the base model, it can be directly adapted to various community models fine-tuned from the base model, as illustrated in \figref{fig:application}(b).

\myPara{Compatible with Spatial-based Add-ons.} ROICtrl is used to separate various instances and can collaborate with fine-grained spatial controls (\eg, ControlNet~\cite{zhang2023adding}, T2I-Adapter~\cite{mou2024t2i}) during inference. As shown in \figref{fig:application}(c), without ROICtrl, T2I-Adapter alone cannot control specific instance and suffers from severe attribute leakage.

\myPara{Compatible with Embedding-based Add-ons.} ROICtrl can work with embedding-based add-ons to control instance identity. As shown in \figref{fig:application}(d, e), it supports both the tuning-based ED-LoRA~\cite{gu2024mix} and the tuning-free IP-Adapter~\cite{ye2023ip}. This compatibility is achieved by reusing pretrained cross-attention without adding new learnable modules for instance captions, ensuring seamless integration with embedding-based add-ons that rely on pretrained cross-attention.

\myPara{Continuous Generation.} ROICtrl enables continuous generation, allowing modification of local regions while preserving previously generated content, as shown in \figref{fig:application}(f).

\begin{figure*}[!tb]
    \centering
\includegraphics[width=0.99\linewidth]{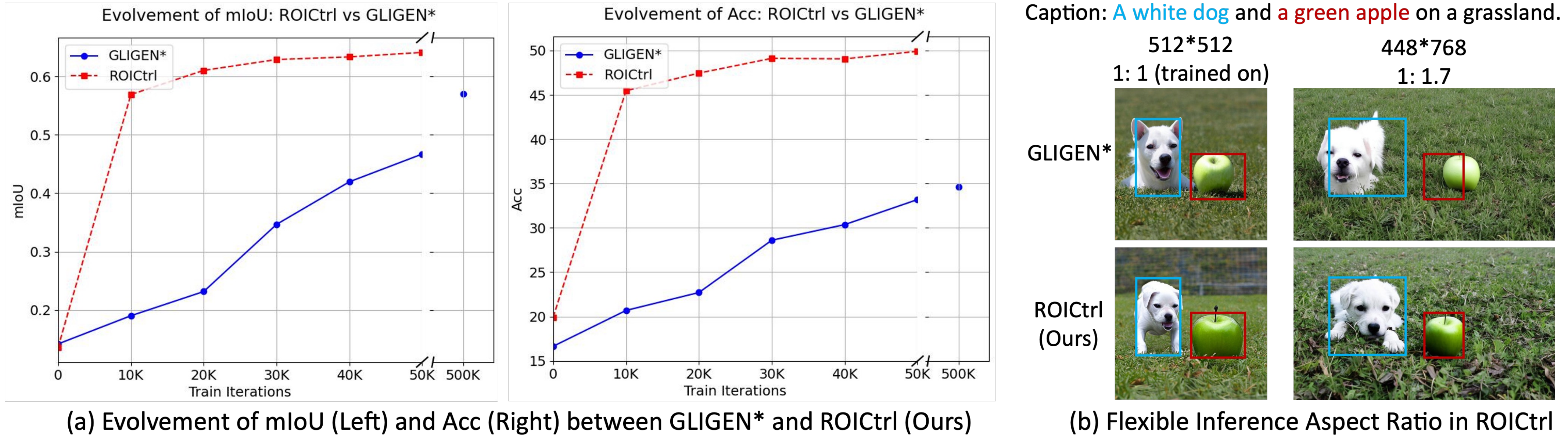}
    \vspace{-.1in}
    \caption{Ablation study comparing ROICtrl and embedding-based injection (GLIGEN*). ROICtrl achieves faster convergence, improved spatial and regional text alignment, and flexible inference aspect ratios.}
\label{fig:ablation_gligen}
    \vspace{-.1in}
\end{figure*}

{
\renewcommand{\arraystretch}{1.2}
\begin{table*}[!tb]
\caption{Ablation study comparing ROICtrl with attention mask–based ROI injection. ROICtrl achieves similar regional text alignment but better spatial alignment, while significantly reducing memory and computational costs. 
The inference speed is tested by generating a 1024$^2$ resolution image with 25 valid ROIs, 50 DDIM~\cite{song2020denoising} steps, and fp16 precision on an A100 GPU.}
\label{tab:ablation_mask}
\vspace{-.15in}
\centering
\resizebox{0.98\linewidth}{!}{\begin{tabular}{l|c|cc|cc|ccc|c|c|c|c}
\toprule
\multicolumn{2}{l|}{\multirow{2}{*}{\textbf{Models}}} &
  \multicolumn{2}{c|}{\textbf{ROICtrl-Bench}} &
  \multicolumn{2}{c|}{\textbf{MIG-Bench}} &
  \multicolumn{3}{c|}{\textbf{Instdiff-Bench}} &
  \textbf{Training} &
  \textbf{Inference}&
  \textbf{Deployed} &
  \textbf{Support} \\
\multicolumn{2}{l|}{} &
  mIoU &
  Acc &
  mIoU &
  Acc &
  AP &
  Color Acc &
  Texture Acc &
  \textbf{Memory (G)} &
  \textbf{Speed (s/img)} &
  \textbf{Resolution} &
  \textbf{Emb Addon} \\ \midrule
\multicolumn{2}{l|}{ROICtrl (Ours)} & 0.652 & 48.7 & 0.66 & 0.73 & 41.0 & 62.3 & 29.3 & 34.3 & 13.1 & all & \cmark \\\midrule
\multirow{3}{*}{\rotatebox[origin=c]{90}{Mask-Attn}} &
  ROICtrl (mask) & 0.628 & 49.2 & 0.64 & 0.71 & 37.1 & 62.5 & 30.3 & 65.5 & 31.5 & all & \cmark\\
 &
  Instance Diffusion & 0.607
   & 45.6
   & 0.46
   & 0.51
   & 40.1
   & 55.2
   & 26.1
   & -
   & 129.2
   & all
   & \xmark
   \\
 &
  MIGC & 0.490 & 38.9 & 0.56 & 0.66 & 22.4 & 53.8 & 24.3 & - & 23.5 & 8$\times$, 16$\times$ & \xmark\\\bottomrule
  
\end{tabular}}
\vspace{-.18in}
\end{table*}}
\section{Experiments}
\subsection{Experimental Setup}
We implement ROICtrl using the PyTorch~\cite{paszke2019pytorch} framework. For enhanced computational efficiency, we develop a custom CUDA kernel for the ROI-Unpool operation.
We adopt multi-scale ROI, where the ROI size $r \times r$ is defined by the relation $r = 6 \times \log_2 R - 11$, with $R \times R$ representing the spatial feature size. For example, if the diffusion model operates at a resolution of $512 \times 512$, with feature resolutions $R = \{64, 32, 16, 8\}$, the corresponding ROI sizes are $r = \{25, 19, 13, 7\}$.
The model is trained on the MS-COCO~\cite{lin2014microsoft} training set, where we recaptioning each instance with free-form text generated by CogVLM~\cite{betker2023improving}. Training is performed with a batch size of 128 and a learning rate of 5e-5, over 60,000 steps on 8 NVIDIA A100 GPUs.

\subsection{Evaluation Setup}
\myPara{ROICtrl-Bench.} 
Existing benchmarks for instance control primarily evaluate template-based instance captions, as shown in \tabref{tab:bench_comp}. For example, MIG-Bench~\cite{zhou2024migc} primarily uses templates such as ``\texttt{[adj.]-colored-[noun.]}" for evaluation. However, in real-world applications, users require free-form instance captions to capture a broader range of attributes. To bridge this gap, we construct ROICtrl-Bench, which consists of four tracks to evaluate various settings. Tracks 1 and 2 examine template-based instance captions, while Tracks 3 and 4 evaluate free-form instance captions. Additionally, Tracks 2 and 4 assess the model's ability to generate out-of-distribution instance captions generated by GPT-4~\cite{achiam2023gpt}.

\myPara{Evaluation Metrics.} Instance control is evaluated on two main aspects: spatial alignment and regional text alignment.
For spatial alignment, we report the Mean Intersection over Union (mIoU). We calculate mIoU by first using Yolo-World~\cite{cheng2024yolo} with a low confidence threshold to detect region proposals based on specified categories. We then apply bipartite matching to pair detected boxes with ground truth boxes and compute the mIoU between them.
For regional text alignment, we input cropped regions of the generated image and their instance captions into MiniCPM-V 2.6~\cite{yao2024minicpm} and using it to evaluate whether they successfully match. The match rate is reported as accuracy (Acc).

For a comprehensive evaluation of ROICtrl, we also adopt two previous benchmarks, MIG-Bench~\cite{zhou2024migc} and InstDiff-Bench~\cite{wang2024instancediffusion}, following their evaluation settings.

\subsection{Comparison to Prior Works}

\myPara{Quantitative Comparison.}
We present the quantitative comparison results in \tabref{tab:quantitative}. Notably, ROICtrl outperforms both Instance Diffusion~\cite{wang2024instancediffusion} and MIGC~\cite{zhou2024migc} in generating small objects, as illustrated in \tabref{tab:quantitative}(b). This improvement is due to ROICtrl's precise localization of small objects, which avoids the quantization errors commonly introduced by masked attention.
On ROICtrl-Bench (\tabref{tab:quantitative}(c)), ROICtrl performs slightly worse than Instance Diffusion on out-of-distribution subjects. This discrepancy may be due to Instance Diffusion being trained on an internal dataset containing 5 million recaptioned images, whereas our approach is trained solely on the publicly available MS-COCO dataset~\cite{lin2014microsoft} with 118K samples. Despite this, \tabref{tab:quantitative} clearly shows that ROICtrl outperforms previous methods, achieving superior spatial alignment and regional text alignment.

\myPara{Qualitative Comparison.}
As qualitative comparison summarized in \figref{fig:qualitative_comp}(a), ROICtrl effectively models occlusions when bounding boxes overlap. In \figref{fig:qualitative_comp}(b), when encountering out-of-distribution instance captions, ROICtrl shows fewer attribute leakage compared to GLIGEN~\cite{li2023gligen} and MIGC~\cite{zhou2024migc}, while maintaining better global consistency than Instance Diffusion~\cite{wang2024instancediffusion}. Additionally, in \figref{fig:qualitative_comp}(c) and (d), ROICtrl achieves superior regional text alignment with free-form instance caption compared to previous methods.

\subsection{Ablation Study}

\subsubsection{Compare to ROI-Injection via Embedding}

We compare ROICtrl with embedding-based ROI injection (see \figref{fig:roicomp}(a)). We adopt the architecture from GLIGEN~\cite{li2023gligen}, and label this variant as GLIGEN*. As summarized in~\figref{fig:ablation_gligen}(a), ROICtrl consistently outperforms GLIGEN* at same training steps. Notably, even after 500K training steps (10$\times$ more than ROICtrl), GLIGEN* still exhibits weaker spatial alignment.
In terms of regional text alignment, GLIGEN* uses global self-attention to inject instance captions, leading to severe attribute leakage and much poorer regional text alignment than ROICtrl.

Additionally, embedding-based ROI injection struggles to generalize when the inference aspect ratio deviates from the training aspect ratio, as shown in~\figref{fig:ablation_gligen}(b), which poses challenges for practical applications that require flexible inference aspect ratio.

\begin{table}[!tb]
\caption{Ablation study of design choices in ROICtrl.}
\label{tab:ablation_design}
\vspace{-.12in}
\centering
\resizebox{\linewidth}{!}{\begin{tabular}{l|cc|cc|ccc}
\toprule
\multirow{2}{*}{\textbf{Models}} &
  \multicolumn{2}{c|}{\textbf{ROICtrl-Bench}} &
  \multicolumn{2}{c|}{\textbf{MIG-Bench}} &
  \multicolumn{3}{c}{\textbf{Instdiff-Bench}}\\
  & mIoU & Acc & mIoU & Acc & AP & Color Acc & Texture Acc \\ \midrule
  ROICtrl (Ours) & 0.652 & 48.7 & 0.66 & 0.73 & 41.0 & 62.3 & 29.3\\\midrule
   $-$ ROI Self-Attn & 0.540 & 48.6 & 0.66 & 0.72 & 32.7 & 60.5 & 32.9\\
   $-$ $\mathcal{L}_{reg}$
   & 0.658 & 47.2 & 0.66 & 0.72 & 41.1 & 58.2 & 21.9 \\
   global coord $\rightarrow$ local coord & 0.655 & 49.5 & 0.68 & 0.74 & 42.1
   & 63.3 & 30.3 \\
   multi-scale roi $\rightarrow$ single-scale roi & 0.639 & 49.6 & 0.65 & 0.73 & 40.0 & 62.5 & 29.9\\
   \bottomrule
  
\end{tabular}}
\vspace{-.1in}
\end{table}

\begin{figure}[!tb]
    \centering
\includegraphics[width=\linewidth]{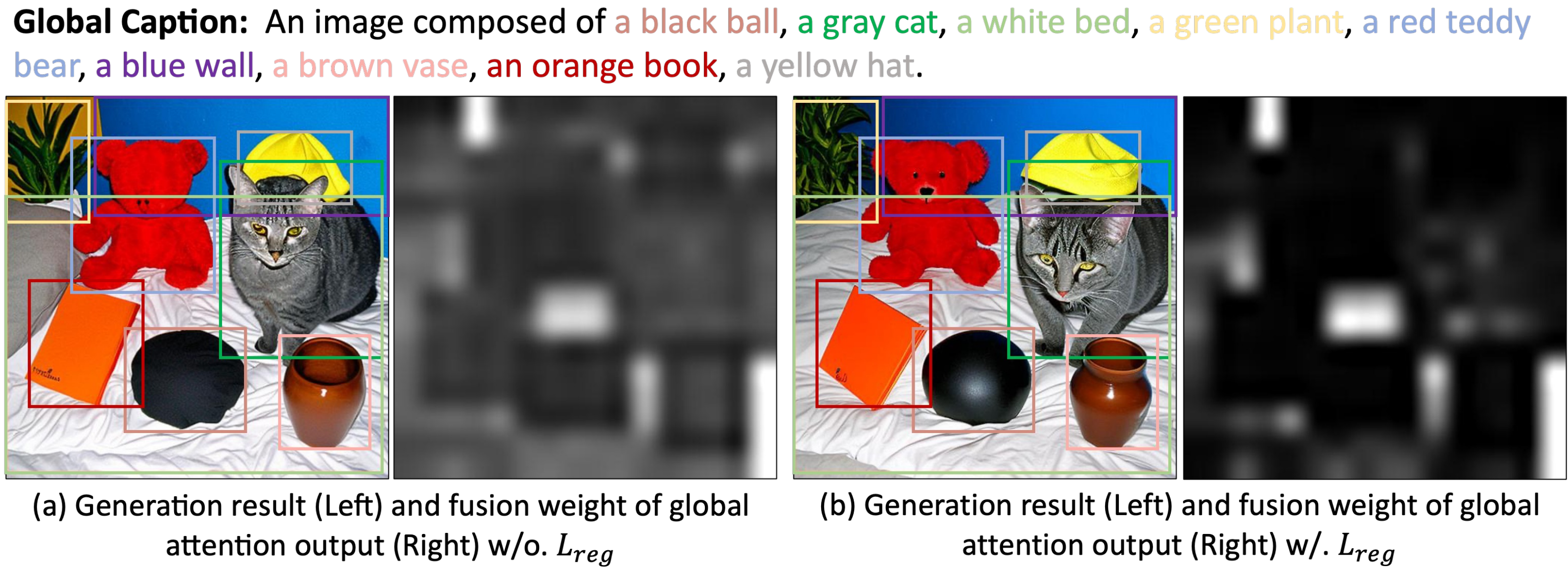}
    \vspace{-.27in}
    \caption{Effect of global attention regularization $\mathcal{L}_{reg}$. Adding $\mathcal{L}_{reg}$ reduces the weight of the global attention output within the ROI, leading to improved regional text alignment.}
\label{fig:ablation_reg}
    \vspace{-.1in}
\end{figure}

\subsubsection{Compare to ROI-Injection via Attention Mask}

We compare ROICtrl with attention mask–based ROI injection (see \figref{fig:roicomp}(b)). Specifically, we modify the ROICtrl implementation by replacing ROI-Align and ROI-Unpool with masked attention. This variant, labeled as ROICtrl (Mask), achieves similar regional text alignment but much worse spatial alignment than ROICtrl, while also consuming more training memory and reducing inference speed, as summarized in \tabref{tab:ablation_mask}.

We also compare ROICtrl with previous attention mask-based methods, Instance Diffusion~\cite{wang2024instancediffusion} and MIGC~\cite{zhou2024migc}. Instance Diffusion is more computationally intensive due to its complex design for supporting point and mask control, with ROICtrl achieving about 9.8$\times$ speedup. MIGC reduces computation by deploying adapters only on low-resolution feature maps (\eg, 8$\times$ or 16$\times$ downsampled feature), resulting in degraded spatial and regional text alignment performance, while still being slower than ROICtrl. Additionally, both Instance Diffusion and MIGC rely on learnable modules for regional caption injection, making them incompatible with embedding-based add-ons like ED-LoRA~\cite{gu2024mix} and IP-Adapter~\cite{ye2023ip}.

\subsubsection{Key Design Choices of ROICtrl}
\myPara{Effect of ROI Self-Attn.} In ROICtrl, we reuse pretrained cross-attention to inject instance captions. However, since this cross-attention is initially trained on full-resolution features, applying it directly to ROI features without ROI self-attention refinement leads to poorer spatial alignment. As shown in \tabref{tab:ablation_design}, removing ROI self-attention decreases the mIoU on ROICtrl-Bench from 0.652 to 0.540 and the AP on InstDiff-Bench from 41.0 to 32.7.

\myPara{Effect of $\mathcal{L}_{reg}$.} As discussed in \secref{sec:objective}, $\mathcal{L}_{reg}$ is used to decrease the fusion weight of the global attention output within the ROI. As shown in \tabref{tab:ablation_design}, omitting $\mathcal{L}_{reg}$ reduces the impact of the instance caption, results in poorer regional text alignment: Color Acc decreases from 62.3 to 58.2 and Texture Acc drops from 29.3 to 21.9 on InstDiff-Bench. As visualization shown in \figref{fig:ablation_reg}, applying $\mathcal{L}_{reg}$ enhances the instance textures of the ball and teddy bear.

\begin{figure}[!tb]
\centering
\includegraphics[width=\linewidth]{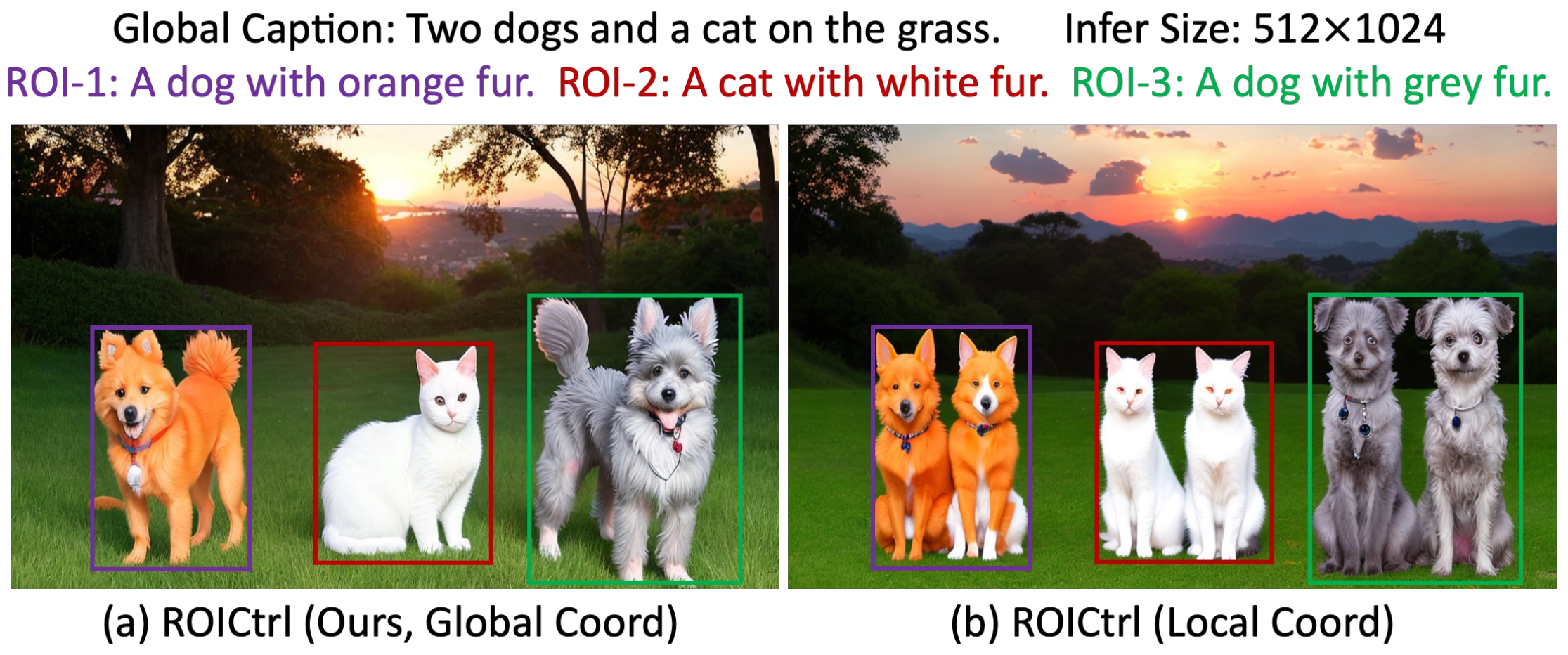}
  \vspace{-.25in}
  \caption{Comparison of regional and global coordinate conditioning. Regional coordinate conditioning leads to repetition issues when the inference size is doubled relative to the training size.}
  \label{fig:ablation_scale}
    \vspace{-.25in}
\end{figure}

\myPara{Regional \textit{vs.} Global Coordinate Conditioning.} 
ROICtrl employs global coordinate conditioning, following GLIGEN~\cite{li2023gligen}, whereas recent works such as MIGC~\cite{zhou2024migc} and BlobGEN~\cite{nie2024compositional} utilize regional coordinate conditioning. \tabref{tab:ablation_design} shows that local coordinate conditioning achieves slightly better quantitative performance. However, in real-world applications, we find that local coordinate conditioning does not generalize well to varying resolutions. As shown in \figref{fig:ablation_scale}, with an inference size of 512$\times$1024 (double the training size of 512$\times$512), regional coordinate conditioning suffers from subject repetition issues.

\myPara{Multi-Scale ROIs.}
In ROICtrl, we set multi-scale ROIs to adapt to the multi-scale feature maps of U-Net. We compare the multi-scale ROIs with single-scale ROIs, where $r = \{7, 7, 7, 7\}$, as shown in \tabref{tab:ablation_design}. Multi-scale ROIs achieve better spatial alignment while maintaining similar text alignment.

\section{Conclusion}
In this paper, we introduce ROICtrl, a method designed to boost instance control in visual generation. ROICtrl is built upon ROI-Unpool, a foundational operation that enables efficient ROI modeling on high-resolution feature maps. By leveraging ROICtrl, we adapt existing diffusion models and their add-ons for multi-instance generation, showcasing a variety of applications enabled by our approach. ROICtrl demonstrates superior qualitative and quantitative results across various benchmarks while also achieving notable speedup, paving the way for controllable generation of multi-instance compositions.

{
\small
\bibliographystyle{ieeenat_fullname}
\bibliography{main}
}

\cleardoublepage
\section{Detailed Evaluation Settings}
\subsection{ROICtrl-Bench}
ROICtrl-Bench contains 200 samples, divided into groups \{1, 2, 3, 4, 5, 6-10, 11-15, 16-20, 21-25, 26-30\} based on instance counts. Each group includes 20 examples randomly selected from the MS-COCO 2017 evaluation set~\cite{lin2014microsoft}. Half of the evaluation examples contain small-sized ROIs with spatial size smaller than $32\times 32$.

As discussed in Sec. \textcolor{red}{4.2}, we create four types of instance captions for each example, corresponding to four tracks, resulting in a total of 800 evaluation examples. For template-based captions in tracks 1 and 2, we follow the GLIGEN~\cite{li2023gligen} evaluation protocol, using only category labels as instance captions. For free-form instance captions, we leverage a multi-modal large language model to provide instance captions. We report the spatial alignment (mIoU) and regional text alignment (Acc) metrics for each track.

\begin{figure}[!tb]
    \centering
    \includegraphics[width=\linewidth]{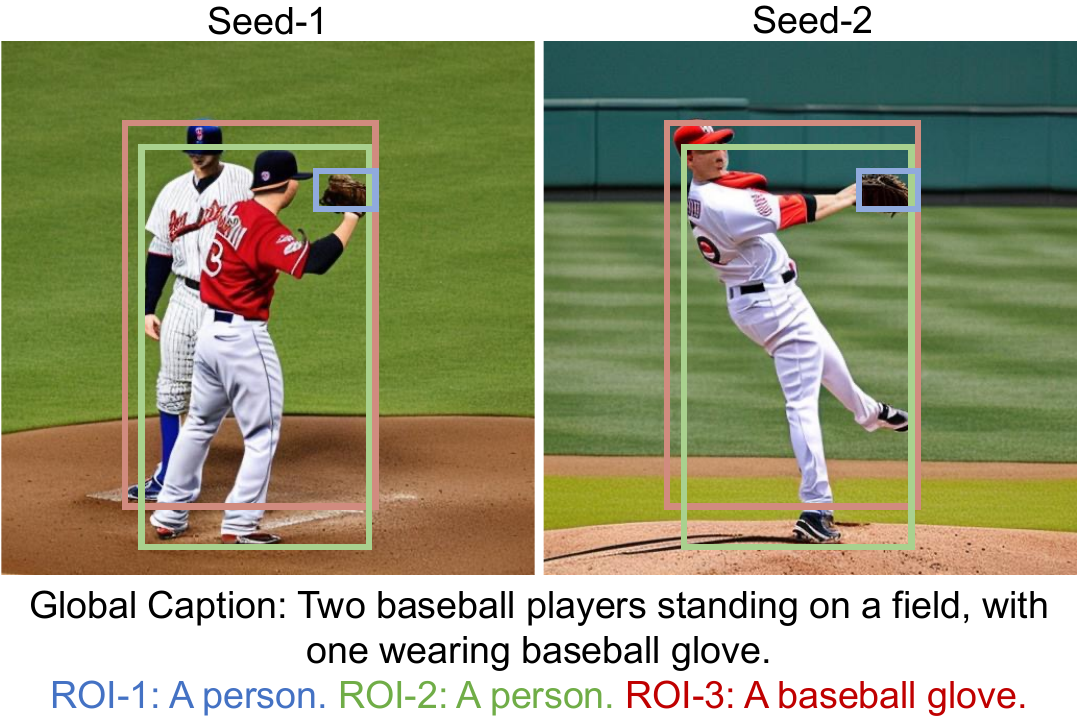}
    \vspace{-.25in}
    \caption{Limitation of ROICtrl. ROICtrl prioritizes the use of instance captions to solve attribute binding but performs unstably when instance boxes with similar captions are heavily overlapped.}
    \label{fig:limitation}
\end{figure}

\subsection{InstDiff-Bench}
InstDiff-Bench~\cite{wang2024instancediffusion} uses the entire MS-COCO 2017 evaluation set~\cite{lin2014microsoft} as its benchmark. For spatial alignment evaluation, it calculates YOLOv8 detection metrics (AP) based on in-distribution instance captions (\ie, object categories). To assess the model's ability to generate out-of-distribution instance captions, it defines 8 common colors: {black, white, red, green, yellow, blue, pink, purple}, and 8 common textures: {rubber, fluffy, metallic, wooden, plastic, fabric, leather, glass}. For each instance, a texture or color adjective is randomly selected from that predefined adjective pool, and the caption is constructed using the template \texttt{[adj.]-[noun.]}.
InstDiff-Bench inputs the cropped box into the CLIP model to predict attributes (colors and textures) and evaluates the accuracy of the predicted adjectives (\ie, $\text{Acc}_{\text{color}}$ or $\text{Acc}_{\text{texture}}$). Additionally, it reports the regional CLIP score for each instance caption.

\begin{figure*}[!tb]
    \centering
    \begin{subfigure}[b]{\linewidth}
        \centering
        \includegraphics[width=\linewidth]{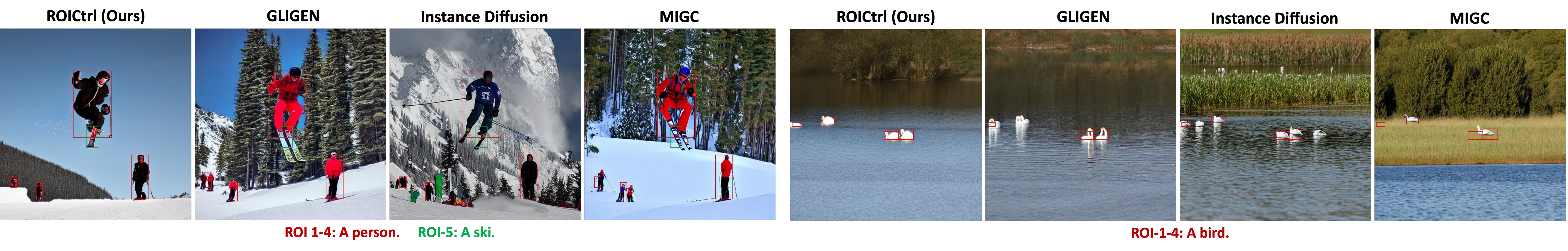}
        \caption{Qualitative comparison of ROICtrl and previous methods on \textit{\textbf{small-sized ROIs}} in Instdiff-Bench~\cite{wang2024instancediffusion}. (\textbf{\textit{Zoom in}} for details.)}
    \end{subfigure}
    \hfill
    \begin{subfigure}[b]{\linewidth}
        \centering
        \includegraphics[width=\linewidth]{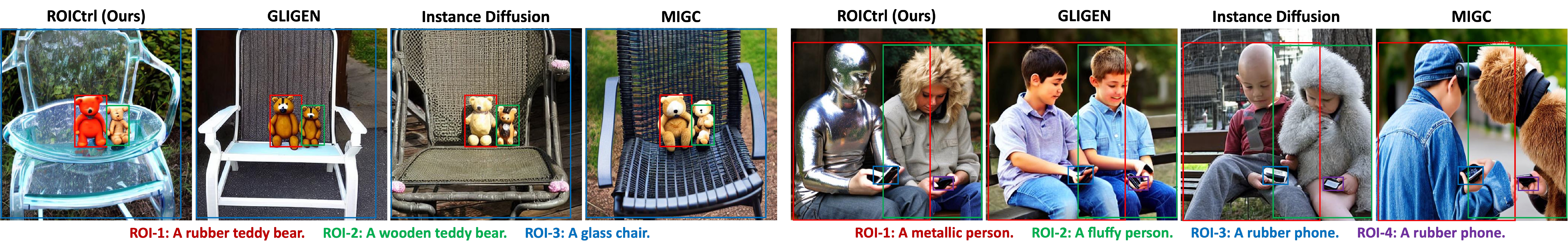}
        \caption{Qualitative comparison of ROICtrl and previous methods on \textit{\textbf{out-of-distribution instance captions}} in Instdiff-Bench~\cite{wang2024instancediffusion}.}
    \end{subfigure}
    \hfill
    \begin{subfigure}[b]{\linewidth}
        \centering
        \includegraphics[width=\linewidth]{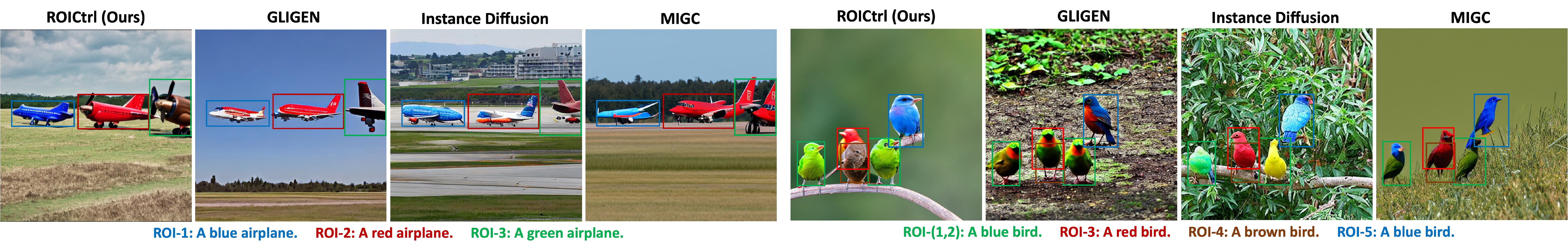}
        \caption{Qualitative comparison of ROICtrl and previous methods on \textit{\textbf{out-of-distribution instance captions}} in MIG-Bench~\cite{zhou2024migc}.}
    \end{subfigure}
    \caption{Qualitative comparison of ROICtrl and previous methods on Instdiff-Bench~\cite{wang2024instancediffusion} and MIG-Bench~\cite{zhou2024migc}. We provide examples for small-sized ROIs and out-of-distribution instance captions.}
    \label{fig:qualitative_instdiff_mig}
\end{figure*}

\subsection{MIG-Bench}
MIG-bench~\cite{zhou2024migc} mainly evaluates spatial alignment and regional text alignment on out-of-distribution instance captions. It selects 800 layouts from COCO, randomly assigns a color to each instance, and constructs the caption based on the template \texttt{[adj.]-colored-[noun]}. In their evaluation, they filter out small-sized ROIs and dense ROIs with more than 6 instances. MIG-bench primarily reports spatial alignment (mIoU) and regional text alignment (instance success rate).

\section{Additional Experiments}
\subsection{Qualitative Comparison}
We have demonstrated the qualitative comparison on ROICtrl-Bench in Sec. \textcolor{red}{4.3} of the main paper. Therefore, in this section, we primarily present the qualitative comparison on InstDiff-Bench~\cite{wang2024instancediffusion} and MIG-Bench~\cite{zhou2024migc}.

\myPara{Small-Sized ROIs.}
As shown in \figref{fig:qualitative_instdiff_mig}(a), previous instance diffusion~\cite{wang2024instancediffusion} tends to generate redundant instances beyond the box, while GLIGEN~\cite{li2023gligen} and MIGC~\cite{zhou2024migc} do not accurately follow the box. In comparison, ROICtrl can accurately generate small-sized ROIs.

\myPara{Out-of-Distribution Instance Caption.}
As shown in \figref{fig:qualitative_instdiff_mig}(b, c), previous methods do not accurately follow the instance caption when generating out-of-distribution attributes and exhibit attribute leakage. In comparison, ROICtrl  follows the instance caption accurately.

\begin{figure*}[!tb]
    \centering
    \animategraphics[width=\linewidth,loop]{8}{images/video_applications/}{00001}{00016}
    \vspace{-.3in}
    \caption{Applications of ROICtrl on Video Instance Control. We encourage readers to \textcolor{magenta}{click and play} the video clips in this figure using Adobe Acrobat.}
    \label{fig:videoapp}
    \vspace{-.1in}
\end{figure*}

\section{Limitation and Future Works}
\subsection{Limitation Analysis}
The attribution leakage problem is largely addressed in ROICtrl, as we prioritize using instance captions in the learnable blending process. However, generating the same instance for highly overlapping bounding boxes remains a challenge. As illustrated in \figref{fig:limitation}, when the boxes exhibit significant overlap, the model need to rely on the global caption for additional information. However, ROICtrl tends to favor instance captions instead, making it unstable to solve this case. We believe that further improving the learnable blending strategy to dynamically reweight the global and instance captions could solve this issue.

\subsection{Future Works}
\myPara{Apply ROICtrl to Video Instance Control.} In our preliminary experiments on VideoCrafter2~\cite{chen2024videocrafter2}, we find that with slight fine-tuning of the pretrained ROICtrl on a video dataset (about 2K iterations), ROICtrl can be used to control video instances, as shown in \figref{fig:videoapp}. However, improving the temporal consistency of video instances remains a challenge, presenting a potential direction for future development of ROICtrl.

\myPara{Apply ROI-Unpool to Diffusion Transformers.} ROICtrl is primarily designed for UNet-based diffusion models. Another future direction is to explore combining ROI-Unpool with transformer-based diffusion models~\cite{esser2024scaling, peebles2023scalable} to explicitly separate instance features and inject instance control.

\subsection{Potential Negative Social Impact}
This project aims to provide the community with an effective method for performing multi-instance control.
However, a risk exists wherein malicious entities could exploit this framework, in combination with image customization, to generate deceptive images of multiple public figures, potentially misleading the public. This concern is not owing to our approach but rather a shared consideration in concept customization. One potential solution to mitigate such risks involves adopting methods similar to anti-dreambooth~\cite{van2023anti}, which introduce subtle noise perturbations to the published images to mislead the customization process. Additionally, applying unseen watermarking to the generated image could deter misuse and prevent them from being used without proper recognition.

\end{document}